\documentclass[letterpaper]{article} 
\usepackage{aaai25}  
\usepackage{times}  
\usepackage{helvet}  
\usepackage{courier}  
\usepackage[hyphens]{url}  
\usepackage{graphicx} 
\usepackage{booktabs}
\usepackage{amssymb}
\usepackage{subfigure}
\usepackage{amsmath}
\newcommand{\correspondingauthor}{\textsuperscript{*}}
\usepackage{natbib}  
\usepackage{caption} 
\usepackage{algorithm}
\usepackage{algorithmic}

\usepackage{newfloat}
\usepackage{listings}
\DeclareCaptionStyle{ruled}{labelfont=normalfont,labelsep=colon,strut=off} 
\floatstyle{ruled}
\newfloat{listing}{tb}{lst}{}
\floatname{listing}{Listing}
\title{Achieving More with Less: A Tensor-Optimization-Powered Ensemble Method}
\author{
    Jinghui Yuan\textsuperscript{\rm 1,\rm 2}, Weijin Jiang\textsuperscript{\rm 3}, Zhe Cao\textsuperscript{\rm 1}, Fangyuan Xie\textsuperscript{\rm 1,\rm 2}\\Rong Wang\textsuperscript{\rm 1}\correspondingauthor, Feiping Nie\textsuperscript{\rm 1,\rm 2}\thanks{Corresponding author}, Yuan Yuan\textsuperscript{\rm 1}
}
\affiliations{
    \textsuperscript{\rm 1}School of Artificial Intelligence, OPtics and ElectroNics (iOPEN),\\
 Northwestern Polytechnical University, Xi’an, 710072, Shaanxi, P. R. China\\
    \textsuperscript{\rm 2}Institute of Artificial Intelligence (TeleAI), China Telecom, P. R. China \\
    \textsuperscript{\rm 3}School of Automation, Northwestern Polytechnical University, Xi'an, 710072, Shaanxi, P. R. China\\
    yuanjh@mail.nwpu.edu.cn, weijinjiang0@gmail.com, caozhe@mail.nwpu.edu.cn, xiefangyuan@mail.nwpu.edu.cn\\ wangrong07@tsinghua.org.cn, feipingnie@gmail.com, y.yuan1.ieee@gmail.com
}

\usepackage{bibentry}

\begin{document}

\maketitle

\begin{abstract}
Ensemble learning is a method that leverages weak learners to produce a strong learner. However, obtaining a large number of base learners requires substantial time and computational resources. Therefore, it is meaningful to study how to achieve the performance typically obtained with many base learners using only a few. We argue that to achieve this, it is essential to enhance both classification performance and generalization ability during the ensemble process. To increase model accuracy, each weak base learner needs to be more efficiently integrated. It is observed that different base learners exhibit varying levels of accuracy in predicting different classes. To capitalize on this, we introduce confidence tensors $\tilde{\mathbf{\Theta}}$, where $\tilde{\mathbf{\Theta}}_{rst}$ signifies the degree of confidence that the $t$-th base classifier assigns the sample to class $r$ while it actually belongs to class $s$. To the best of our knowledge, this is the first time an evaluation of the performance of base classifiers across different classes has been proposed. The proposed confidence tensor compensates for the strengths and weaknesses of each base classifier in different classes, enabling the method to achieve superior results with a smaller number of base learners. To enhance generalization performance, we design a smooth and partially convex objective function that leverages the concept of margin, making the strong learner more discriminative. Furthermore, it is proved that in the gradient matrix of the loss function, the sum of each column's elements is zero, allowing us to solve a constrained optimization problem using gradient-based methods. We then compare our algorithm with random forests of ten times the size and other classical methods across numerous datasets, demonstrating the superiority of our approach. Finally, we discuss the reasons for the success of our algorithm and point out that it can be naturally extended from bagging to stacking.
\end{abstract}

%

\section{Introduction}
Ensemble learning is a method that constructs a strong learner from weak learners. To date, ensemble learning has achieved success in many fields. \cite{zhou2019deep} apply the concept of ensemble learning to deep ensemble methods. \cite{liu2015spectral} combine deep learning with spectral clustering. \cite{wood2023unified} attempt to explain ensemble learning from the diversity perspective. \cite{shi2022multi} consider multi-view ensemble learning. \cite{cao2020ensemble} introduce ensemble learning into bioinformatics for interdisciplinary research. In summary, ensemble learning is a very important concept in the field of artificial intelligence.

However, ensemble learning often requires integrating a large number of base learners. For example, \cite{chen2016xgboost} design XGBoost using the boosting idea, and \cite{sun2024improved} consider reducing the correlation among random forests to improve their performance. Nevertheless, all these ensemble methods have relatively high time complexity. It is worthwhile to investigate how to achieve the effects of a large ensemble using only a few base learners.

We believe that to achieve this goal, it is necessary to adjust the confidences during the ensemble process. On one hand, the accuracy of the combined classifiers should be as high as possible, and on the other hand, generalization should be as good as possible. 

To increase the accuracy of ensemble model, we need each base learner to be integrated more efficiently. We observe that different base learners have varying accuracy for different classes. For example, one learner may be very accurate in classifying classes $1$ and $2$ but less accurate for class $3$, while another learner may be accurate for classes $2$ and $3$ but less accurate for class $1$. Based on this, we propose a learnable confidence tensor $\tilde{\mathbf{\Theta}}$, where $\tilde{\mathbf{\Theta}}_{rst}$ represents the probability that the $t$-th base classifier classifies an instance as class $r$ when it actually belongs to class $s$. This concept is analogous to the confusion matrix \cite{townsend1971theoretical}. To better represent the tensor, we use a technique to unfold the tensor and prove that it has favorable properties, which can assist us in solving the constrained optimization problem \cite{boyd2004convex}.

Another important issue is how to improve the generalization of the ensemble. We introduce the concept of margin into multi-class optimization because \cite{zhou2014large} points out that margin is a key factor in suppressing overfitting in AdaBoost. Additionally, \cite{nie2024multi} achieve good results by considering margin in multi-class support vector machines. \cite{chen2023incremental} propose the OMD theory using the margin concept, demonstrating excellent performance across multiple datasets.

There has been much discussion on how to introduce the margin. For instance, \cite{liu2024nonlinear} defined the margin using the $\ell_{0/1}$ norm, and \cite{li2021beyond} proposed the class margin. However, the margins obtained by these methods are often non-smooth or non-convex, which makes them relatively difficult to optimize \cite{khaled2023unified}. 

To solve the problem, we use the logsumexp \cite{calafiore2019log} technique to derive a loss function with smoothness and certain convexity. Not only that, but we also prove that our loss function satisfies the property of having a gradient column sum of zero, which allows for adaptively adding confidences to each base learner given a sufficiently good initialization, which can be efficiently solved using gradient-based methods \cite{zhao2024adaptivity}.

We compare the proposed algorithm with a random forest that has ten times the number of parameters \cite{biau2016random}, as well as with classical boosting algorithms and support vector machine methods. Experiments demonstrate that we have achieved our goal, outperforming a large number of base learners using only a few.

Finally, we conduct an in-depth discussion, analyzing the relationship between our algorithm and dropout \cite{liu2023dropout} and parameter freezing \cite{wimmer2023dimensionality}. We discuss and analyze why our algorithm performs better. Additionally, we point out that this algorithm can not only be used as bagging \cite{ngo2022evolutionary} but can also be naturally extended to stacking \cite{zounemat2021ensemble}.

We summarize our main contributions below:
\begin{itemize}
\item We propose a learnable tensor $\tilde{\mathbf{\Theta}}$, which encapsulates the confidences of each base learner for different classes. We design a loss function based on the margin concept, which is smooth and partially convex.
\item We prove that the loss function has the desirable property of having a gradient column sum of zero. This allows us to solve the proposed optimization problem with linear constraints using gradient-based methods, and we design an algorithm for this purpose.
\item We conduct extensive comparative experiments. Under the same conditions, the ensemble of a small number of base learners outperforms a random forest with ten times the number of base learners, as well as other classical algorithms, validating our method's superiority.
\item We further discuss the relationship between our algorithm and random forests, explaining that random forests can be seen as a result of our algorithm after applying dropout. We also elucidate why our algorithm performs well and how it can be naturally extended to stacking methods.

\end{itemize}

\section{Methodology}
In this section, we will introduce the symbols used in the paper, the construction of the loss function, and the specific optimization model employed.
\subsection{Notations}
Assuming we are addressing a $c$-class classification problem, where 
$x_i$ represents the $i$-th sample before ensemble training with $k$ classifiers $G_1,...,G_k$, each basic classifier inputs $x_i$
and outputs a c-dimensional indicator column vector $G_j(x_i)=(0,...,1,...,0)^T\in \mathbb{R}^{c \times 1}$. Suppose the true labels are represented by $y \in \mathbb{R}^{c\times1}$, and each element of $y$ is one-hot encoded, where $m_i$ is the class of the $i$-th sample, forming the indicator matrix $Y \in \mathbb{R}^{c\times n}$. If $g_i$ is a column vector satisfying $g_i=(G_1^T(x_i),...,G_k^T(x_i))^T$,  then $g_i\in \mathbb{R}^{kc\times 1}$. Combining all of $g_i$ yields the matrix $G=(g_1,...,g_n)$, where matrix $G\in \mathbb{R}^{kc\times n}$.

\subsection{Introduction of Margin}
Margin represents the distance from data points to the decision boundary. First, let's introduce how predictions are made using the confidence tensor $\tilde{\mathbf{\Theta}}$. The tensor $\tilde{\mathbf{\Theta}}$ has three dimensions: $c$ dimensions, $c$ dimensions, and $k$ dimensions. Here, $\tilde{\mathbf{\Theta}}_{rst}$ signifies that the $t$-th base classifier assigns the sample to class $r$ while it actually belongs to class $s$ . The probabilities for all classes are combined into this confidence tensor $\tilde{\mathbf{\Theta}}$. However, tensors are not straightforward to handle in matrix computations. Therefore, we split it into matrix $\Theta$ using forward slicing, as illustrated in the diagram below, so $\Theta \in \mathbb{R}^{c\times kc}$.
\begin{figure}[h]
\centering
 \includegraphics[width=0.46\textwidth]{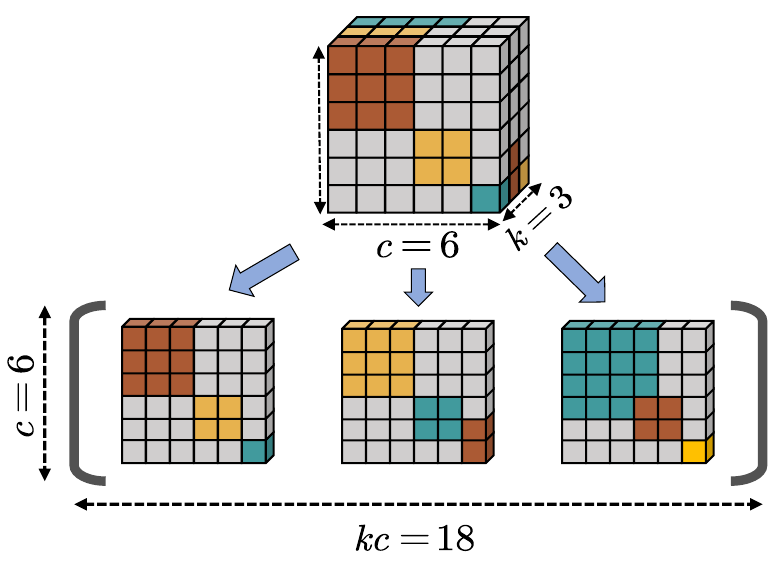}
	\caption{Expansion diagram of $\tilde{\mathbf{\Theta}}$}
	\label{F6}
 \end{figure}
 
Define $\mathcal{S}$ as the softmax function. When $\mathcal{S}$ is applied to a column vector, it performs softmax normalization on the column vector. When $\mathcal{S}$ is applied to a matrix, it applies the softmax function to each column of the matrix. Based on this, the prediction for the $i$-th sample can be expressed as
\begin{equation}
    \hat{y}_i=\underset{j = 1...c}{argmax}\ \left(\mathcal{S}\left(\Theta g_i\right)\right)_j
\end{equation}

A suitable definition of the margin for multi-class classification is the difference between the predicted value at the true label position and the second largest predicted value. This can be expressed as formula 
\begin{equation}
Y_i^T(Y_i\odot\mathcal{S}(\Theta g_i))-max_2(\mathcal{S}\left(\Theta g_i\right))
\end{equation}
with the symbol $max_2(v)$ representing the second largest element in vector $v$, $Y_i$ is the $i$-th column of $Y$, additionally, the symbol $\odot$ represents the Hadamard product of matrices. This definition is reasonable as it reflects the confidence in classifying into the true class.

However, both the maximum function and the $max_2$ function are non-smooth, and non-smooth functions can significantly affect the convergence and convergence speed of the algorithm. A reasonable solution is to approximate these functions using the log-sum-exp function. For a vector $v$, by choosing a suitable $\alpha$, the log-sum-exp function can effectively approximate the largest element in $v$. The expression for the log-sum-exp function is given by Eq.(3).
\begin{equation}
f(v)=\frac{1}{\alpha}log\left(\sum_{j=1}^{c}e^{\alpha v_j}\right)
\end{equation}

The choice of parameter $\alpha$ is straightforward and does not significantly affect the degree of approximation \cite{logsumexp}. Through extensive experiments, we found that when $\alpha$ is set to 10, it can already extract the maximum value with considerable accuracy.

Using the log-sum-exp function, the $max_2$ function can be easily represented. To extract the second largest value, we only need to set the maximum position to 0 and then retrieve the maximum value again. We assume that the predicted value at the true class is the largest, which is reasonable, and we will explain this later. This can be expressed using the following formula:
\begin{equation}
max_2(\mathcal{S}\left(\Theta g_i\right))=\frac{1}{\alpha}log\left(\sum_{j=1}^{c}e^{\alpha (\mathcal{S}\left(\Theta g_i\right)-Y_i\odot\mathcal{S}\left(\Theta g_i\right))_j}\right)
\end{equation}

Thus, the margin contributed by the $i$-th item can be expressed as Eq.(5),
\begin{equation}
Y_i^T(Y_i\odot\mathcal{S}\left(\Theta g_i\right))-\frac{1}{\alpha}log\left(\sum_{j=1}^{c}e^{\alpha (\mathcal{S}\left(\Theta g_i\right)-Y_i\odot\mathcal{S}\left(\Theta g_i\right))_j}\right)
\end{equation}
and the total margin can be expressed as:
\begin{equation}
\begin{aligned}
&\mathcal{M}=\sum_{i=1}^n
 Y_i^T(Y_i\odot\mathcal{S}\left(\Theta g_i\right))\\
&-\sum_{i=1}^n\frac{1}{\alpha}log\left(\sum_{j=1}^{c}e^{\alpha (\mathcal{S}\left(\Theta g_i\right)-Y_i\odot\mathcal{S}\left(\Theta g_i\right))_j}\right)
\end{aligned}
\end{equation}

\subsection{Introduction of Loss Function}
Now let's return to consider the assumptions made at Eq.(4). In the previous margin calculation, the actual $max_2(\mathcal{S}\left(\Theta g_i\right))$ should be the maximum of 
\begin{equation}
    \mathcal{S}\left(\Theta g_i\right)-Y^\mu\odot max(\mathcal{S}\left(\Theta g_i\right)) 
\end{equation}
where $\mu$ is $\underset{j=1...c}{argmax}(\mathcal{S}\left(\Theta g_i\right))_j$. $Y^\mu$ represents the column vector where the $\mu$-th element is 1 and others are 0. But we calculate the maximum of 
\begin{equation}
    \mathcal{S}\left(\Theta g_i\right)-Y_i\odot\mathcal{S}\left(\Theta g_i\right)
\end{equation}
this is because subtracting the linear term results in a margin function with better smoothness and convexity. In fact, we make the following assumption:
\begin{equation}
    max(\mathcal{S}\left(\Theta g_i\right)) \approx \| Y_i \odot \mathcal{S}(\Theta g_i) \|
\end{equation}

This essentially assumes correct classification. To account for misclassification, we incorporate a penalty term into the objective function. Cross-entropy can effectively measure accuracy in the classification process. Therefore, we add the accuracy function as follows:
\begin{equation}
\mathcal{C}=-\sum_{i=1}^n(Y_i^T(Y_i\odot log(\mathcal{S}\left(\Theta g_i\right))))
\end{equation}

The final optimization objective is a negative confidence combination of both
\begin{equation}
\mathcal{L}=\mathcal{C}-\gamma \mathcal{M}
\end{equation}
where $\gamma$ is a hyperparameter. A larger $\gamma$ indicates a greater preference for higher confidence in classification, thus a larger margin. Specifically, when $\gamma$ equals 0, it reverts to the standard cross-entropy loss function used for classification. Therefore, the cross-entropy function is a special case of the loss function we are using.

\subsection{Introduction of Optimization Problem}
In this section, we will introduce the final optimization problem. Actually, the accuracy of each base classifier varies, so naturally, the confidences of each base classifier should also differ. Let's denote the accuracy of the $t$-th base classifier as 
$w_t$, where $w=(w_1,...,w_k)^T$, $\tilde{w}$ expands 
$w$ to $kc$ dimensions , expressed in formula as:
\begin{equation}
\tilde{w}=(w_1,...w_1,w_2,...w_2,...,w_k,...,w_k)^T
\end{equation}

Assigning confidences to each classifier can be seen as imposing constraints on $\Theta$'s columns, with the constraint being 
$\Theta^T1=\tilde{w}$.

In summary, the final optimization problem can be formulated as:
\begin{equation}
    \begin{aligned}
        &\underset{\Theta}{min}\ \mathcal{L}\\
        & \Theta^T1=\tilde{w}
    \end{aligned}
\end{equation}
and the loss function $\mathcal{L}$ is
\begin{equation}
    \begin{aligned}
&\mathcal{L}=\mathcal{C}-\gamma\mathcal{M}\\
&=\sum_{i=1}^n-\left(Y_i^Tlog(\mathcal{S}\left(\Theta g_i\right))+\gamma
Y_i^T\mathcal{S}\left(\Theta g_i\right)\right)\\
&+\gamma  \sum_{i=1}^n\frac{1}{\alpha}log\left(\sum_{j=1}^{c}e^{\alpha (\mathcal{S}\left(\Theta g_i\right)-Y_i\odot\mathcal{S}\left(\Theta g_i\right))_j}\right)   
    \end{aligned}
\end{equation}

\section{Optimization Algorithm}
The convexity of the optimization problem is crucial for solving it. Performing the softmax operation inevitably transforms the convex function into a non-convex one. Fortunately, apart from the softmax operation, our problem remains convex. In other words, we have the following theorem.
\subsubsection{Theorem 1.}The loss function $\mathcal{L}$ is a convex function with respect to $\mathcal{S}(\Theta g_i)$.
\subsubsection{Proof} Assume that matrix $D$ has only one non-zero element $D_{m_i m_i}=1$, then $Y_i\odot\mathcal{S}(\Theta g_i)$ can be seen as a linear mapping of $\mathcal{S}(\Theta g_i)$, i.e. $D\mathcal{S}(\Theta g_i)$, and so that 
\begin{equation}
   \mathcal{S}(\Theta g_i)-Y_i\odot\mathcal{S}(\Theta g_i) = (I-D)\mathcal{S}(\Theta g_i) 
\end{equation}

As is well-known, for a convex function $f(x)$, $f(Ax+b)$ is also a convex function. The log-sum-exp function is a well-known convex function, and the other two terms are evidently convex as well. The non-negative sum of convex functions is convex. Thus, the proof is complete.

Excellent convexity allows us to use gradient methods to quickly reach a local minimum. The key to solving the above optimization problem is how to handle constraint $\Theta^T1=\tilde{w}$. Through the following theorem, we can naturally satisfy the constraint with a simple initialization method.

\subsubsection{Theorem 2.}Assume $\nabla_{kl}\mathcal{L}$ represents the kl-th element of the gradient of $\mathcal{L}$ with respect to matrix $\Theta$ i.e. $\frac{\partial \mathcal{L}}{\partial \Theta}_{kl}$,  then we have the following formula $\sum_{k=1}^c\nabla_{kl}\mathcal{L}=0$.
\subsubsection{Proof}Without loss of generality, we assume there is only one data, which belongs to the $m$-th class. That corresponds to the $m$-th row of $Y$ being 1 and all other rows being 0. Therefore, there is no need for summation, and the loss function can be expressed as:
\begin{equation}
    \begin{aligned}
&\mathcal{L}=\mathcal{L}_1+\mathcal{L}_2+\mathcal{L}_3\\
&= - Y^Tlog(\mathcal{S}\left(\Theta g\right)) - \gamma
Y^T\mathcal{S}\left(\Theta g\right)\\
&+\gamma\frac{1}{\alpha}log\left(\sum_{j=1}^{c}e^{\alpha (\mathcal{S}\left(\Theta g\right)-Y\odot\mathcal{S}\left(\Theta g\right))_j}\right)   
    \end{aligned}
\end{equation}

Suppose $\delta_{ij}$ represents the Kronecker delta function, which is 1 when $i=j$ and 0 otherwise.

According to the chain rule, it is easy to verify that the partial derivative of the j-th component of $\mathcal{S}(\Theta g)$ with respect to  $\Theta_{kl}$ satisfies:
\begin{equation}
    \begin{aligned}
\frac{\partial \mathcal{S}(\Theta g)_j }{\partial \Theta_{kl}} =   \mathcal{S}(\Theta g)_j(g_l\delta_{jk}-\mathcal{S}(\Theta g)_kg_l)
    \end{aligned}
\end{equation}

Therefore, according to Eq.(17) the derivative of the first term $\mathcal{L}_1=- Y^Tlog(\mathcal{S}\left(\Theta g\right))$ with respect to $\Theta_{kl}$ is easily known to be
\begin{equation}
    \begin{aligned}
&\frac{\partial \mathcal{L}_1 }{\partial \Theta_{kl}} =-\frac{\partial log(\mathcal{S}(\Theta g))_m }{\partial \Theta_{kl}}\\
&=-\frac{1}{\mathcal{S}(\Theta g)_m}\mathcal{S}(\Theta g)_m(g_l\delta_{mk}-\mathcal{S}(\Theta g)_kg_l)\\
&=-(g_l\delta_{mk}-\mathcal{S}(\Theta g)_kg_l)
    \end{aligned}
\end{equation}
Similarly, the partial derivative of the second term with respect to $\Theta_{kl}$ is 
\begin{equation}
    \begin{aligned}
\frac{\partial \mathcal{L}_2 }{\partial \Theta_{kl}}= -\gamma\mathcal{S}(\Theta g)_m(g_l\delta_{mk}-\mathcal{S}(\Theta g)_kg_l)
    \end{aligned}
\end{equation}

Noting formula $\sum_{k=1}^c(g_l\delta_{mk}-\mathcal{S}(\Theta g)_kg_l)=0$, we have $\sum_{k=1}^c\nabla_{kl}(\mathcal{L}_1+\mathcal{L}_2)=0$. Next, we only need to prove that the derivative of the third term with respect to $k$ sums to zero. It is easy to verify that the derivative of the third term can be expressed in the following form:
\begin{equation}
    \begin{aligned}
\frac{\partial \mathcal{L}_3 }{\partial \Theta_{kl}} &=\gamma\frac{\sum_{j=1}^ce^{\alpha\mathcal{S}(\Theta g)_j}\mathcal{S}(\Theta g)_j(g_l\delta_{jk}-\mathcal{S}(\Theta g)_kg_l)}{\sum_{j=1}^ce^{\alpha\mathcal{S}(\Theta g)_j}-e^{\alpha\mathcal{S}(\Theta g)_m}+1}\\
&-\gamma\frac{e^{\alpha\mathcal{S}(\Theta g)_m}\mathcal{S}(\Theta g)_m(g_l\delta_{mk}-\mathcal{S}(\Theta g)_kg_l)}{\sum_{j=1}^ce^{\alpha\mathcal{S}(\Theta g)_j}-e^{\alpha\mathcal{S}(\Theta g)_m}+1}
    \end{aligned}
\end{equation}

Since Eq.(20) also includes formula $(g_l\delta_{mk}-\mathcal{S}(\Theta g)_kg_l)$ and we know that $\sum_{k=1}^c(g_l\delta_{mk}-\mathcal{S}(\Theta g)_kg_l)=0$, then we have $\sum_{k=1}^c\nabla_{kl}\mathcal{L}_3=0$. 

Thus, we have 
\begin{equation}
    \begin{aligned}
\sum_{k=1}^c\nabla_{kl}\mathcal{L}=\sum_{k=1}^c\nabla_{kl}(\mathcal{L}_1+\mathcal{L}_2+\mathcal{L}_3)=0
    \end{aligned}
\end{equation}
it means that we have proven this theorem.

It is worth noting that Eq.(18) (19) (20) not only help us prove this theorem, but also directly provide the formula for the gradient. This allows us to compute the gradient directly using these three formulas and apply gradient descent to optimize the objective function. The theorem proven above is a very useful property that aids us in quickly obtaining the optimal solution for the constrained loss function.

In fact, for non-special cases, i.e. with more than one sample, suppose set $\{\mathcal{M}\}$ represents formula $\{m_1,...,m_n\} $, where $m_i\in \{1,...,c\}$. Then the final gradient is equal to
\begin{equation}
    \begin{aligned}
&\frac{\partial \mathcal{L} }{\partial \Theta_{kl}} = -\sum_{i=1}^n(1+\gamma\mathcal{S}(\Theta g)_{m_i})(g_l\delta_{m_ik}-\mathcal{S}(\Theta g)_kg_l)\\
&+\sum_{i=1}^n\left(\gamma\frac{\sum_{j=1}^ce^{\alpha\mathcal{S}(\Theta g)_j}\mathcal{S}(\Theta g)_j(g_l\delta_{jk}-\mathcal{S}(\Theta g)_kg_l)}{\sum_{j=1}^ce^{\alpha\mathcal{S}(\Theta g)_j}-e^{\alpha\mathcal{S}(\Theta g)_{m_i}}+1}\right)\\
&-\sum_{i=1}^n\left(\gamma\frac{e^{\alpha\mathcal{S}(\Theta g)_{m_i}}\mathcal{S}(\Theta g)_{m_i}(g_l\delta_{m_ik}-\mathcal{S}(\Theta g)_kg_l)}{\sum_{j=1}^ce^{\alpha\mathcal{S}(\Theta g)_j}-e^{\alpha\mathcal{S}(\Theta g)_{m_i}}+1} \right)
    \end{aligned}
\end{equation}
where $n$ is the number of data points used to compute the gradient: if using full gradient descent, it includes all data points; if using batch gradient descent, it is the batch size; and if using stochastic gradient descent, it is 1.

Since $\sum_{k=1}^c\nabla_{kl}\mathcal{L}=0$, by using appropriate initialization and gradient-based methods, the constraint can be naturally satisfied. Specifically, the update formula for gradient descent is as follows:
\begin{equation}
    \Theta_{p+1}=\Theta_{p}-\beta\nabla_{\Theta_p}\mathcal{L}(\Theta_p)
\end{equation}
where $\beta$ is the learning rate. If we require that the initialization satisfies $\Theta_01=\tilde{w}$ , then by mathematical induction, it is easy to prove that $\Theta^*1=\tilde{w}$. Therefore, we can initialize each forward slice of the tensor and then expand it as shown below.
\begin{figure}[h]
\centering
 \includegraphics[width=0.46\textwidth]{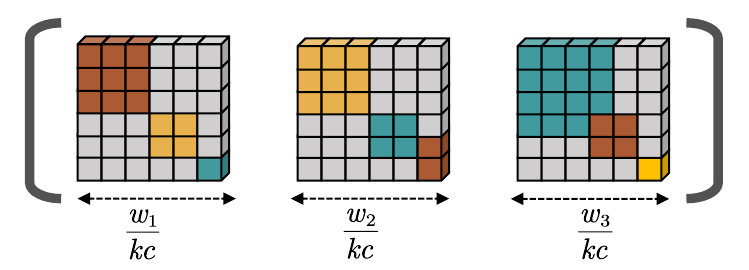}
	\caption{Initialization Diagram}
	\label{F6}
 \end{figure}
 
Due to the partial convexity of our loss function, we can optimize this problem using gradient descent. This involves selecting a batch of $g_i$ at a time to compute the loss function and its gradient. The specific algorithm is illustrated in the flowchart below.

\begin{algorithm}
    \caption{Gradient Descent}
    \label{Algorithm 1} 
    \begin{algorithmic}[1]
        \REQUIRE Matrix $G$
        \ENSURE $\Theta^*$
        \STATE Initialize $\Theta_0$ according to Figure 2.
        \WHILE{not converged}
            \STATE Randomly select a batch of columns $g_i$ from $G$.
            \STATE Compute $\nabla\mathcal{L}(\Theta_p)$ by Eq.(22).
            \STATE Update $\Theta$ by $\Theta_{p+1} = \Theta_p - \beta\nabla_{\Theta}\mathcal{L}(\Theta_p)$.
        \ENDWHILE
    \end{algorithmic}
\end{algorithm}

\subsection{Time Complexity Analysis}
In this section, we will analyze the time complexity of the algorithm. The algorithm computes gradients by selecting a batch of samples at a time. Note that while $g_i$ is a vector of dimension $kc$, according to the definition, $g_i$ is a vector composed of predictions from each classifier, meaning $g_i$ is sparse. Computing the matrix multiplication $\Theta$ with $g_i$ involves adding corresponding positions of $\Theta$ together, without needing multiplication.

In other words, computing the gradient for each element according to Eq.(22) has a time complexity of $\mathcal{O}(kc)$. During the computation, there are many common terms that can be retained to simplify calculations. With $\mathcal{O}(kc^2)$ parameters, the time complexity for updating once will not exceed $\mathcal{O}(k^2c^3)$. Assuming convergence after $N$ iterations, the total time complexity will not exceed $\mathcal{O}(Nk^2c^3)$. 

However, in practice, $k$ and $c$ are usually very small, and there are a lot of direct additions in calculations. Moreover, hardware acceleration with tools like PyTorch enables automatic differentiation, making the actual runtime very fast.

\section{Experiments on toy datasets}
We construct a toy dataset with a double-ring shape, which has the advantage of visualizing classification results. To demonstrate our superiority, we compare with the classical Support Vector Classification (SVC) algorithm and the XGBoost algorithm. To prove that confidence optimization can "achieve more with less", we compare Random Forests with 10 trees (RF10), 20 trees (RF20), 30 trees (RF30), and 100 trees (RF100), while our algorithm used 10 trees (OUR10). All tree models use the same maximum depth. The number of trees in the compared Random Forest models is 1$\times$, 2$\times$, 3$\times$, and 10$\times$ that of our model, respectively.

The results and classification boundaries of the compared algorithms on the toy dataset are shown in Figure 3. Our algorithm used a coefficient $\alpha$ of 10 and $\gamma$ of 5. From the classification boundaries, it can be observed that our algorithm effectively improves the learners' prediction performance near the boundaries.

\begin{figure}[h]
        \centering
        \subfigure[Dataset]
	{
		\includegraphics[width=0.22\textwidth]{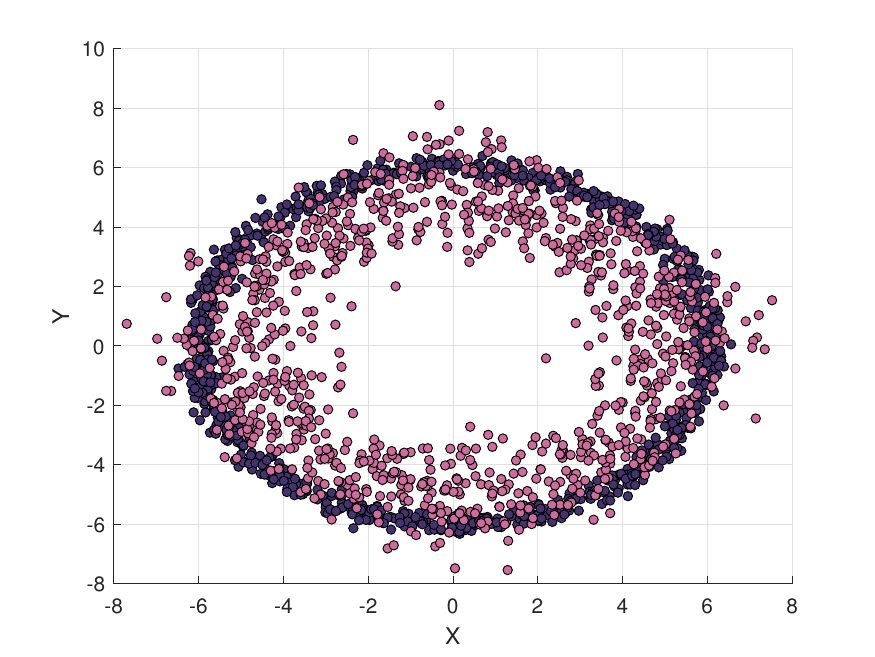}
	}
         \subfigure[OUR10]
	{
		\includegraphics[width=0.22\textwidth]{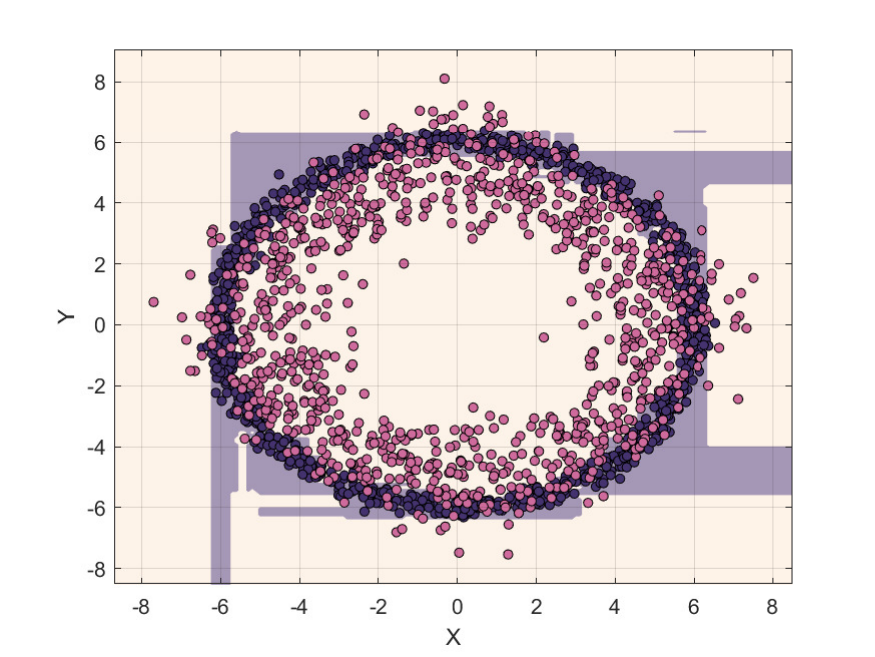}
	}
	\subfigure[SVC]
	{
		\includegraphics[width=0.22\textwidth]{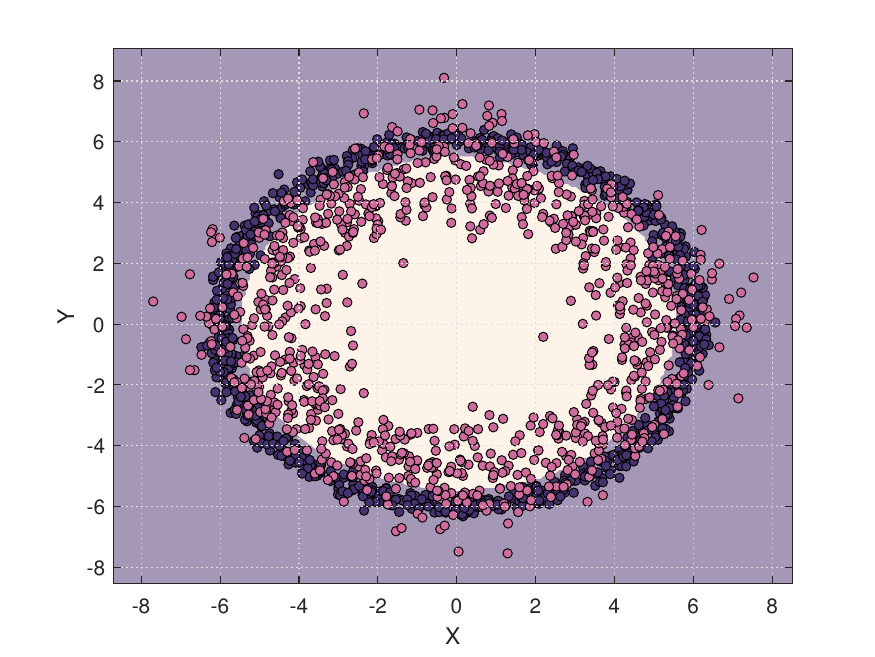}
	}
        \subfigure[XGBoost]
	{
		\includegraphics[width=0.22\textwidth]{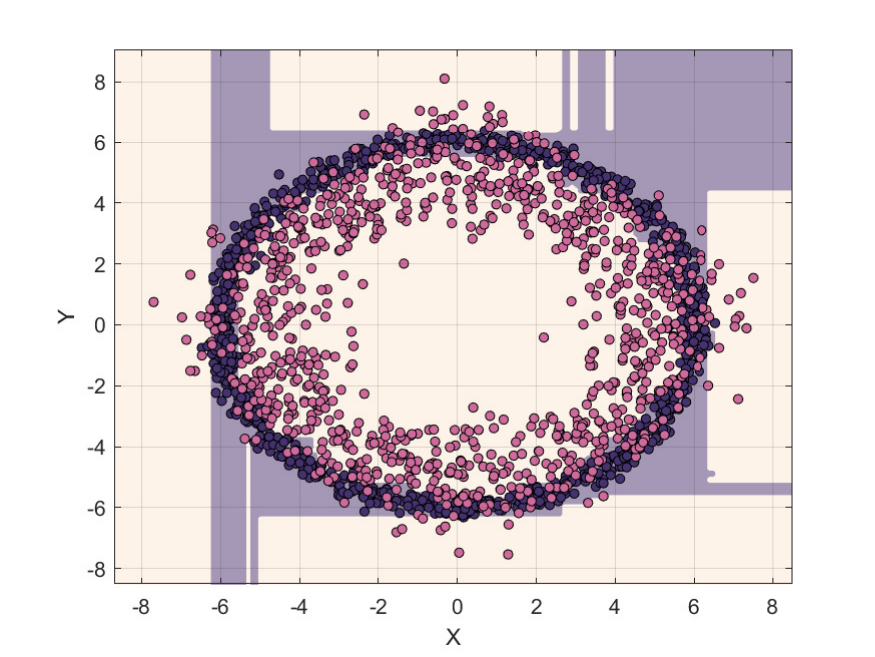}
	}
         \subfigure[RF10]
	{
		\includegraphics[width=0.22\textwidth]{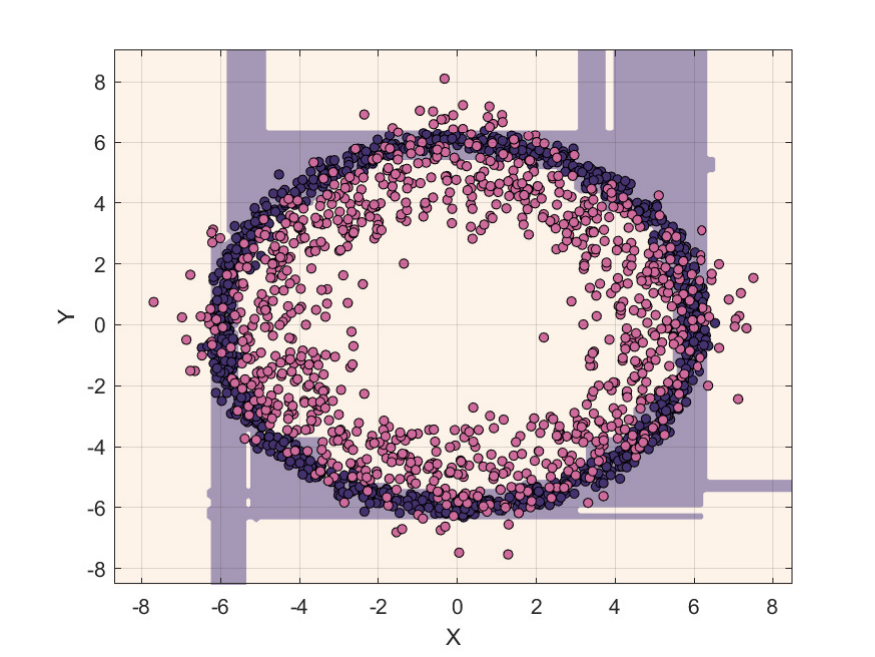}
	}
        \subfigure[RF20]
	{
		\includegraphics[width=0.22\textwidth]{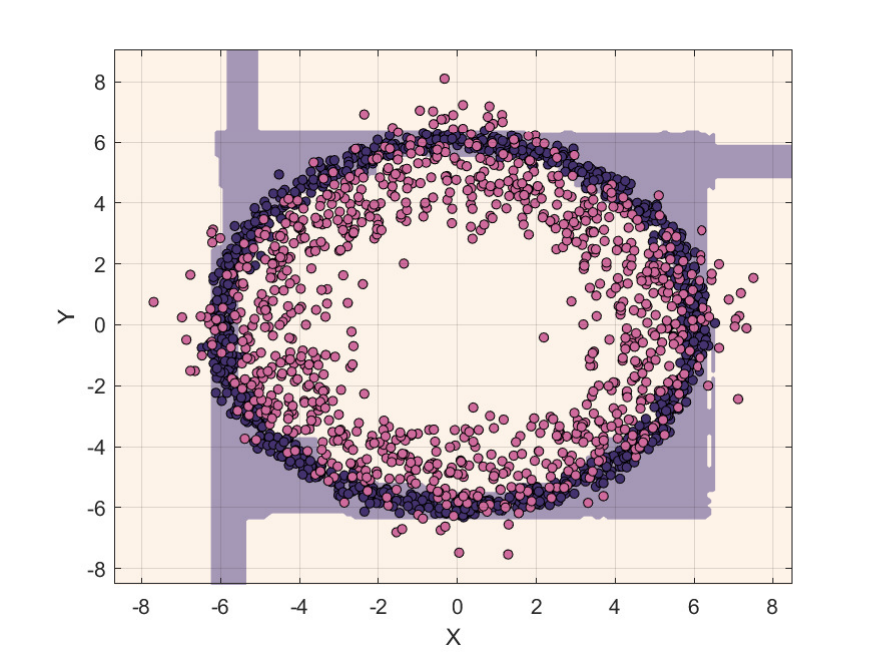}
	}
         \subfigure[RF30]
	{
		\includegraphics[width=0.22\textwidth]{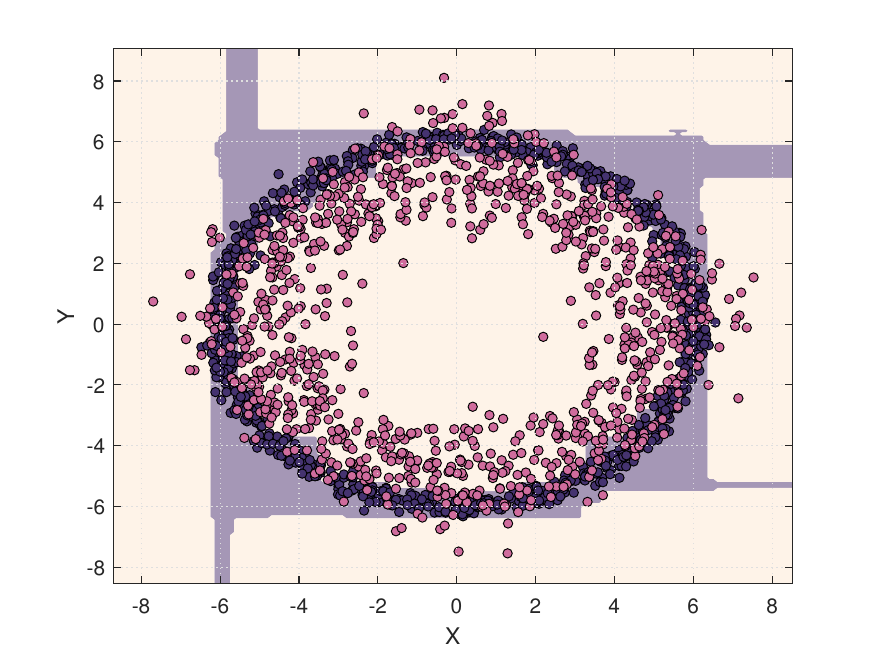}
	}
         \subfigure[RF100]
	{
		\includegraphics[width=0.22\textwidth]{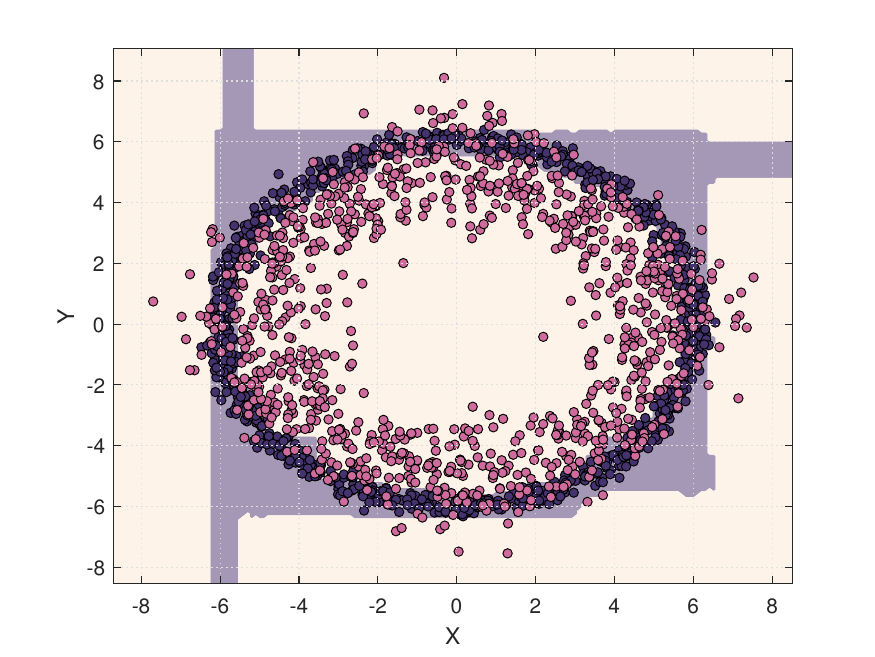}
	}
	\caption{ Performance on toy Dataset. (a) Original dataset. (b) OUR10. (c) SVC. (d) XGBoost. (e) RF10. (f) RF20. (g) RF30. (h) RF100.}
	\label{F6}
\end{figure}

We divide the dataset into training and test sets with an 80\% to 20\% ratio. Below are the performances of the algorithms on the dataset:
\begin{table}[!htbp]
	\centering
	\caption{Results of the toy datasets}
	\label{T1}
	\scalebox{0.8}{
		\begin{tabular}{@{}cccccccc@{}}
			\toprule
			Method     & OUR10 &  SVC & XGBoost & RF10 &RF20 & RF30 &RF100 \\ 
			\midrule
                TrainData	  & 0.902&0.837&\textbf{0.904}&0.883&0.890&0.895&0.899 			\\
                TestData	  & \textbf{0.883}&0.840&0.865&0.863&0.880&0.878&0.880 			\\
                AllData       & \textbf{0.898}&0.838&0.897&0.879&0.888&0.892&0.895     	\\		
			\bottomrule
	\end{tabular}}
\end{table}

From the experimental results on the toy dataset, we can see that although our algorithm does not have the highest accuracy on the training set, it achieves the highest accuracy on both the test set and the entire dataset. This demonstrates that our algorithm effectively suppresses overfitting on the training set and improves overall accuracy, achieving better results with 10 trees compared to the 100-tree Random Forest.

\section{Experiments on real datasets}
To further demonstrate the superiority of our algorithm, we conduct experiments on some real-world datasets of varying sizes. 
\subsection{Experimental Settings}
We conduct extensive experiments on ten datasets, including TR41, warpPIE10P, Movement, Ecoli, Arcene, REUT, GLI 85, Carcinom, Lung and Isolet. Table 2 lists the number of samples and classes for each dataset:
\begin{table}[!htbp]
	\centering
	\caption{Description of the benchmark datasets}
	\label{T1}
	\scalebox{0.9}{
		\begin{tabular}{@{}cccc@{}}
			\toprule
			Datasets     & $\#$Object & $\#$Attribute & $\#$Class \\ 
			\midrule
                TR41	  & 878		& 7454		& 10	\\
               warpPIE10P	  & 210		& 2420		& 10	\\
                Movement   &  360 & 90 & 15\\
                Ecoli & 336 & 7 &8\\
                Arcene &200 &10000 & 2\\
                REUT &10000&2000&4\\
                GLI 85&85&22283&2\\
                Carcinom&174&9782&11\\
                Lung&203&3312&5\\
                Isolet&1560&617&26\\
			\bottomrule
	\end{tabular}}
\end{table}

\subsubsection{Parameters Selection.}
Our algorithm has two hyperparameters, $\alpha$ and $\gamma$. The hyperparameter $\alpha$ is fixed at 10 throughout, and as long as this parameter is not too small, it will not affect the algorithm's performance. The other parameter, $\gamma$, is chosen randomly from $\{5, 10, 15, 20, 25\}$ for each calculation. To facilitate reproducibility, our random seed is fixed at 0.

\subsubsection{Evaluation Metrics.} In our experiments, we always select training set accuracy, test set accuracy, and overall accuracy, with the training set and test set ratio consistently at 80\%:20\%. Our comparison algorithms are the same as those used for the toy dataset, and we always ensure that the basic parameters of the base learners in our algorithm and the Random Forest algorithm are identical.

\subsection{Experimental Results}
Our experimental results are recorded in Tables 3 and 4. Table 3 presents the accuracy of various algorithms on the training set. Table 4 records the accuracy of the algorithms on the test set.

\begin{table}[!htbp]
	\centering
	\caption{Classification accuracy on the training datasets}
	\label{T1}
	\scalebox{0.8}{
		\begin{tabular}{@{}cccccccc@{}}
			\toprule
			Dataset    & OUR10 &  XGBoost & SVC & RF10 &RF20 & RF30 &RF100 \\ 
			\midrule
                Movement	  & \textbf{1.000}&\underline{0.997}&0.903&0.851&0.865&0.899&0.927		\\
                warpPIE10P	  & \textbf{1.000}&1.000&1.000&0.994&1.000&1.000&\underline{1.000} 			\\
                Lung       &\textbf{1.000}&1.000&0.944&0.963&0.994&1.000&1.000     	\\		
                Ecoli	  & \textbf{1.000}&0.963&0.892&0.978&0.993&0.996 &\underline{1.000} 			\\
                Arcene	  & \textbf{1.000}&1.000&0.744&0.994&0.994&1.000&\underline{1.000} 			\\
                Carcinom       & \textbf{1.000}&1.000&0.928&0.993&1.000&\underline{1.000}&1.000    	\\		
                GLI\_85	  & \textbf{1.000}&1.000&0.706&1.000&1.000&1.000&\underline{1.000} 			\\
                Isolet	  & \textbf{1.000}&\underline{1.000}&0.983&0.966&0.984&0.984&0.991 			\\
                TR41       & 0.990&\textbf{1.000}&\underline{0.999}&0.977&0.983&0.976&0.984    	\\	
                REUT      & 0.997&0.929&0.996&0.993&0.998&\underline{0.999}&\textbf{1.000}    	\\	
			\bottomrule
	\end{tabular}}
\end{table}

\begin{table}[!htbp]
	\centering
	\caption{Classification accuracy on the test datasets}
	\label{T1}
	\scalebox{0.8}{
		\begin{tabular}{@{}cccccccc@{}}
			\toprule
			Dataset    & OUR10 &  XGBoost & SVC & RF10 &RF20 & RF30 &RF100 \\ 
			\midrule
                Movement	  & \textbf{0.764}&\underline{0.736}&0.744&0.611&0.569&0.681&0.667		\\
                warpPIE10P	  & \underline{0.976}&0.905&\textbf{1.000}&0.905&0.929&0.929&0.952			\\
                Lung       & \textbf{0.951}&0.927&0.951&0.878&0.951&0.951&\underline{0.951}     	\\		
                Ecoli	  & \textbf{0.882}&\underline{0.868}&0.868&0.853&0.868&0.853 & 0.868			\\
                Arcene	  & \textbf{0.850}&0.650&0.700&0.750&\underline{0.775}&0.750&0.775			\\
                Carcinom       & \textbf{0.886}&0.743&0.686&0.771&0.800&0.857&\underline{0.885}    	\\		
                GLI\_85	  & \textbf{0.941}&0.765&0.647&0.824&0.882&\underline{0.882}&0.882			\\
                Isolet	  & \underline{0.881}&0.878&\textbf{0.955}&0.827&0.881&0.875&0.881 			\\
                TR41       & \textbf{0.972}&\underline{0.972}&0.943&0.960&0.955&0.955&0.955   	\\	
                REUT      & 0.897&0.883&\textbf{0.959}&0.888&0.897&0.900&\underline{0.902}    	\\	
			\bottomrule
	\end{tabular}}
\end{table}

From the experiments, we can see that for the Random Forest algorithm, the larger the number of base classifiers, i.e., decision trees, the higher the accuracy we typically obtain. OUR10 achieves the best performance on several datasets. It even surpasses the Random Forest algorithm with ten times the number of base classifiers than our algorithm, achieving the goal of outperforming large-scale ensemble classifiers with a smaller ensemble.

\subsection{Convergence Analysis}
To verify that our algorithm indeed has very good convergence properties, we select several datasets and plot the iteration curves of our algorithm on these datasets, as shown in Figure 4. We used full gradient descent in the experiments, i.e. computing the gradient using all samples.
\begin{figure}[h]
        \centering
        \subfigure[Movement]
	{
		\includegraphics[width=0.22\textwidth]{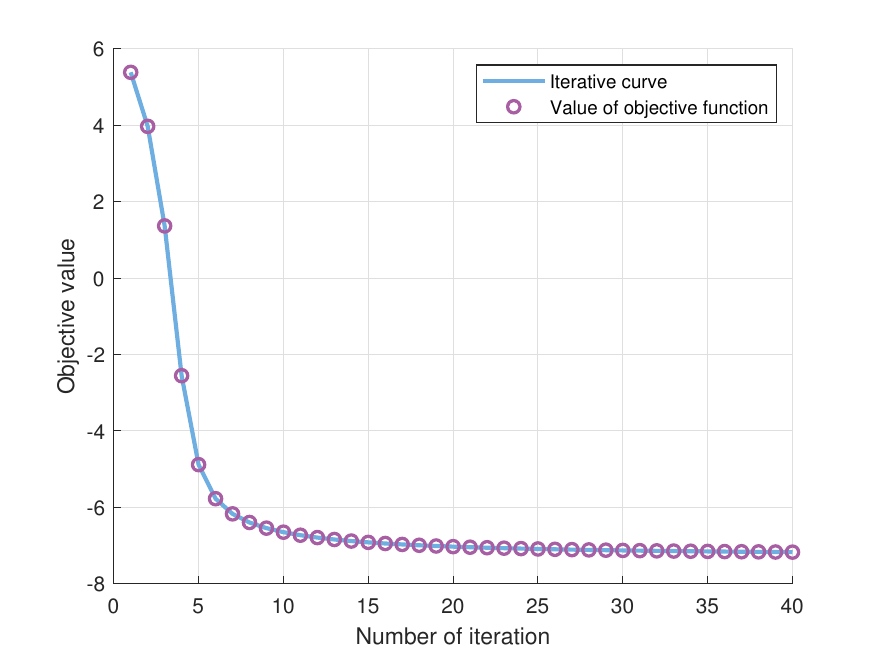}
	}
         \subfigure[TR41]
	{
		\includegraphics[width=0.22\textwidth]{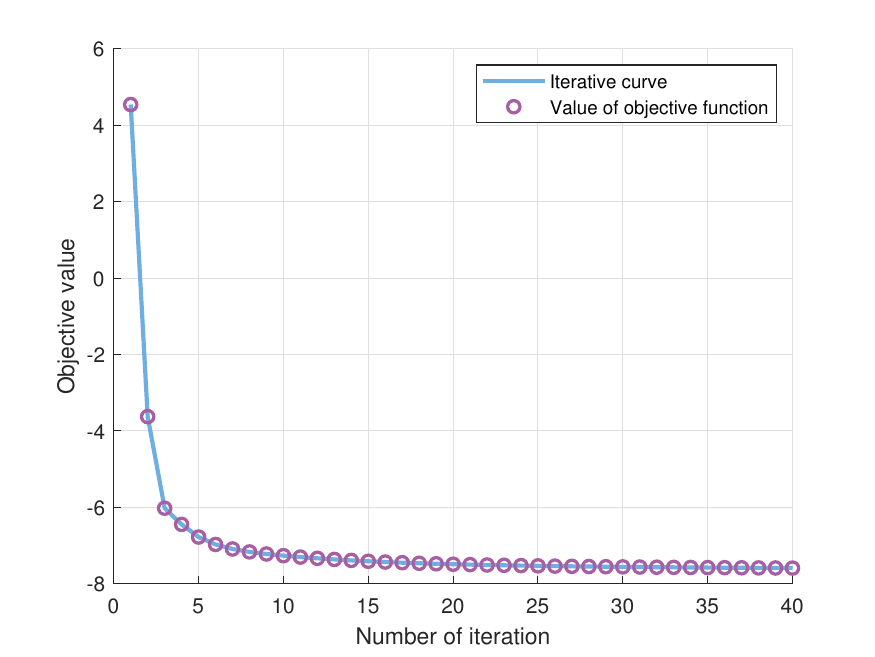}
	}
	\subfigure[warpPIE10P]
	{
		\includegraphics[width=0.22\textwidth]{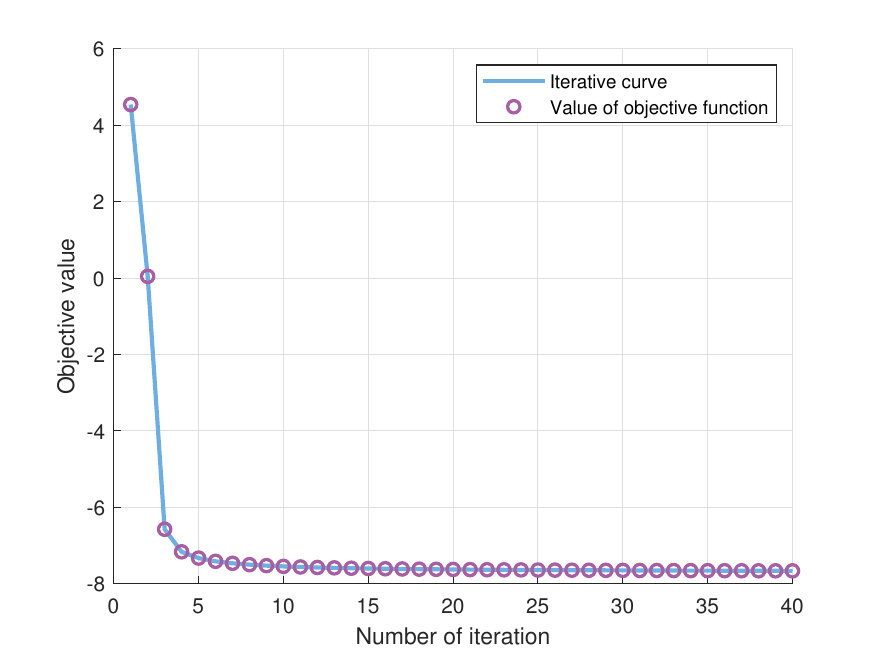}
	}
        \subfigure[Lung]
	{
		\includegraphics[width=0.22\textwidth]{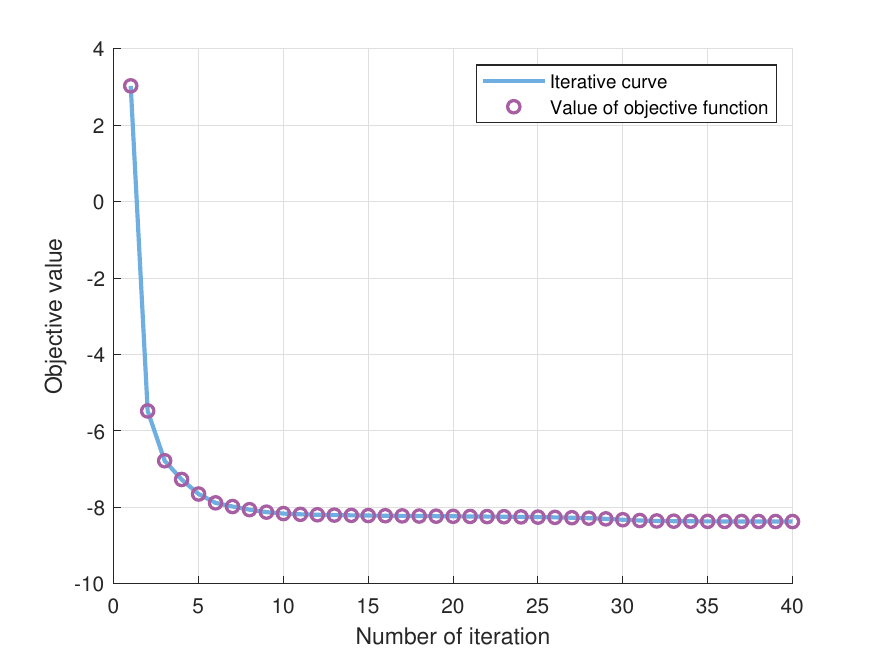}
	}
         \subfigure[Arcene]
	{
		\includegraphics[width=0.22\textwidth]{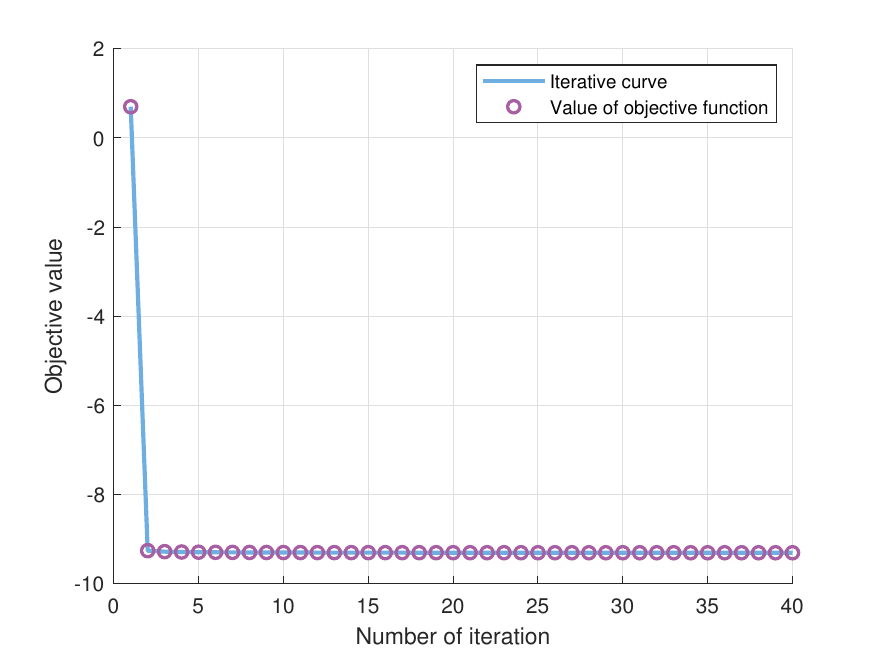}
	}
        \subfigure[Isolet]
	{
		\includegraphics[width=0.22\textwidth]{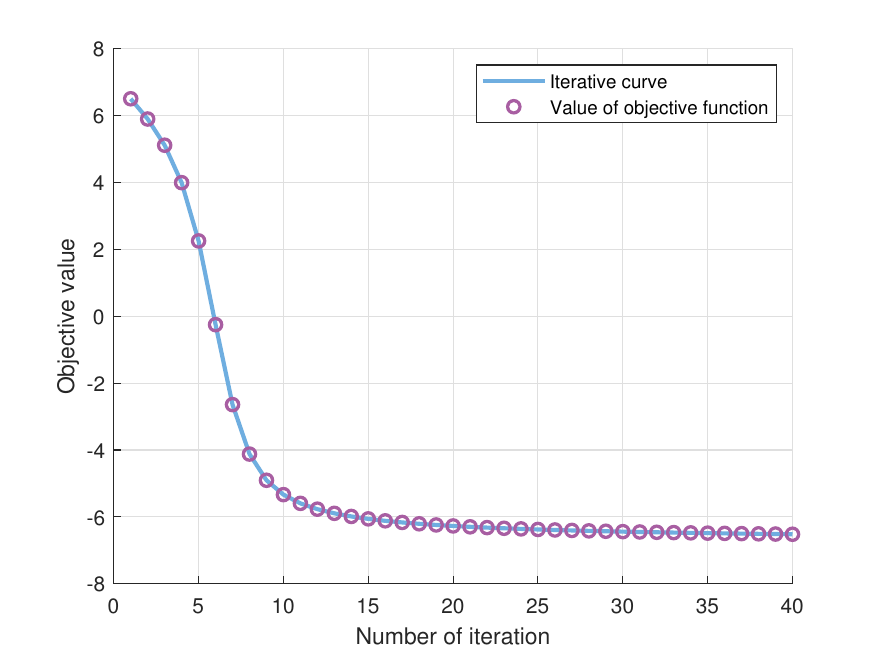}
	}
	\caption{Convergence curve. (a) Movement. (b) TR41. (c) warpPIE10P. (d) Lung. (e) Arcene. (f) Isolet.}
\end{figure}

Our objective function possesses partially convex properties and the algorithm typically converges quickly. As observed in the experiments, convergence is usually achieved within no more than 10 iterations.

\section{Further Discussion}
\subsection{Relationship with Random Forest}
In this section, we will discuss the relationship between our algorithm and the classic random forest, assuming that decision trees are used as base classifiers.
\subsubsection{Dropout} Dropout typically refers to discarding some learned parameters, or setting some parameters to zero. For our learned confidence tensor $\tilde{\mathbf{\Theta}}$, if we apply the dropout idea and drop out all the non-diagonal elements of $\tilde{\mathbf{\Theta}}$, our algorithm would transform into the classic random forest confidence by accuracy.
\subsubsection{Fix Parameters} A more specific case is that after dropping out the non-diagonal elements, we fix the diagonal elements to 1. At this point, each decision tree has the same confidence, and each decision is made by equal-confidence voting. This forms the most classic random forest algorithm.

Since, compared to the classic random forest algorithm, we have more learnable parameters, our algorithm represents an extension of it. 

\subsection{Why Our Algorithm Performs Better}
In this section, we will explain why our algorithm performs better using a simple example.
\subsubsection{What Additional Information We Obtained} Assuming we have learned a confidence matrix $\Theta$, the data it contains is shown in Figure 5. There are three classifiers, and their confidences are 3, 3.6, and 3, respectively. In fact, $\Theta$ is composed of three confusion matrices. We find that the diagonal elements of most matrices are very large, indicating that the classifiers have learned the correct values in most cases, and we trust them more. However, it is worth noting that for the first classifier, the values in the third column are all the same, \textbf{which means the first classifier has not learned how to classify the third category}.

\begin{figure}[h]
\centering
 \includegraphics[width=0.46\textwidth]{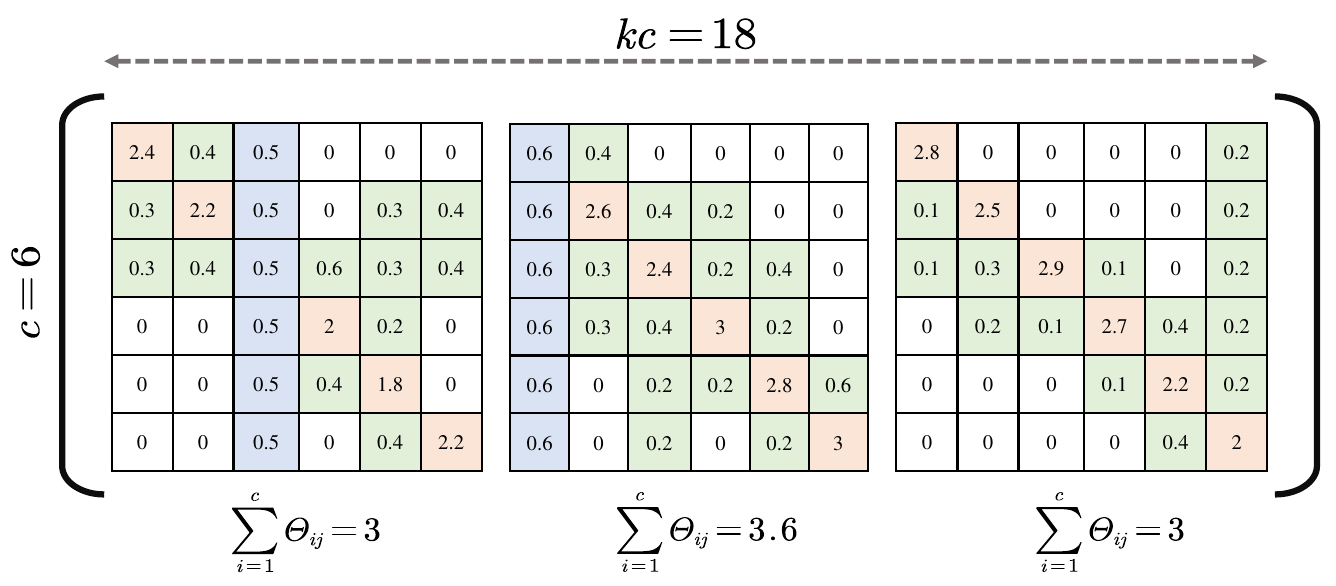}
	\caption{Learned $\Theta$}
 \end{figure}

\subsubsection{confidence Optimization} Note that although the first classifier has not learned to distinguish the third class, it is relatively confident about the other classes. Similarly, the second classifier has not learned to classify the first class but is also confident about the other classes. If using random forest confidence-based voting, the errors introduced by these classifiers cannot be ignored. \textbf{According to our algorithm, the uncertain confidences are very small, which create a complementary relationship with the other confident classifiers, thereby improving accuracy}.

\subsection{From Bagging to Stacking}
Our algorithm seems to learn a set of confidences for bagging, but in fact, it can be easily extended to a stacking algorithm. We do not enforce that base classifiers must be of the same type; instead, we can combine and learn confidences for different base classifiers such as decision trees, support vector machines, and logistic regression. Furthermore, a potential extension is to analyze the performance of heterogeneous base classifiers on different types of datasets.

\section{Conclusion}
In this paper, we focus on how to achieve the performance of using a larger number of base learners through the integration of fewer base learners. To achieve this, we propose that during the integration process of the base learners, we should aim to simultaneously improve both accuracy and generalization. For this purpose, we introduce the concept of margin into the loss function and design a smooth loss function with good convexity using the logsumexp function. We also prove that the column sum of the gradient satisfies good properties, allowing us to solve the constrained optimization problem through gradient descent to obtain the local optimal solution. We compare our method with random forests having 10 times the number of base learners and other classical methods, demonstrating that our integration of fewer learners outperformed the majority. Finally, we discuss the effective reasons behind this algorithm and its further extensions.
\bibliography{aaai25.bib}

@article{zhou2019deep,
  title={Deep forest},
  author={Zhou, Zhi-Hua and Feng, Ji},
  journal={National science review},
  volume={6},
  number={1},
  pages={74--86},
  year={2019},
  publisher={Oxford University Press}
}

@inproceedings{liu2015spectral,
  title={Spectral ensemble clustering},
  author={Liu, Hongfu and Liu, Tongliang and Wu, Junjie and Tao, Dacheng and Fu, Yun},
  booktitle={Proceedings of the 21th ACM SIGKDD international conference on knowledge discovery and data mining},
  pages={715--724},
  year={2015}
}

@article{wood2023unified,
  title={A unified theory of diversity in ensemble learning},
  author={Wood, Danny and Mu, Tingting and Webb, Andrew M and Reeve, Henry WJ and Lujan, Mikel and Brown, Gavin},
  journal={Journal of Machine Learning Research},
  volume={24},
  number={359},
  pages={1--49},
  year={2023}
}

@article{shi2022multi,
  title={When multi-view classification meets ensemble learning},
  author={Shi, Shaojun and Nie, Feiping and Wang, Rong and Li, Xuelong},
  journal={Neurocomputing},
  volume={490},
  pages={17--29},
  year={2022},
  publisher={Elsevier}
}

@article{cao2020ensemble,
  title={Ensemble deep learning in bioinformatics},
  author={Cao, Yue and Geddes, Thomas Andrew and Yang, Jean Yee Hwa and Yang, Pengyi},
  journal={Nature Machine Intelligence},
  volume={2},
  number={9},
  pages={500--508},
  year={2020},
  publisher={Nature Publishing Group UK London}
}

@inproceedings{chen2016xgboost,
  title={Xgboost: A scalable tree boosting system},
  author={Chen, Tianqi and Guestrin, Carlos},
  booktitle={Proceedings of the 22nd acm sigkdd international conference on knowledge discovery and data mining},
  pages={785--794},
  year={2016}
}

@article{sun2024improved,
  title={An improved random forest based on the classification accuracy and correlation measurement of decision trees},
  author={Sun, Zhigang and Wang, Guotao and Li, Pengfei and Wang, Hui and Zhang, Min and Liang, Xiaowen},
  journal={Expert Systems with Applications},
  volume={237},
  pages={121549},
  year={2024},
  publisher={Elsevier}
}

@inproceedings{zhou2014large,
  title={Large margin distribution learning},
  author={Zhou, Zhi-Hua},
  booktitle={Artificial Neural Networks in Pattern Recognition: 6th IAPR TC 3 International Workshop, ANNPR 2014, Montreal, QC, Canada, October 6-8, 2014. Proceedings 6},
  pages={1--11},
  year={2014},
  organization={Springer}
}

@inproceedings{nie2024multi,
  title={Multi-class support vector machine with maximizing minimum margin},
  author={Nie, Feiping and Hao, Zhezheng and Wang, Rong},
  booktitle={Proceedings of the AAAI Conference on Artificial Intelligence},
  volume={38},
  number={13},
  pages={14466--14473},
  year={2024}
}

@inproceedings{chen2023incremental,
  title={Incremental and Decremental Optimal Margin Distribution Learning.},
  author={Chen, Li-Jun and Zhang, Teng and Shi, Xuanhua and Jin, Hai},
  booktitle={IJCAI},
  pages={3523--3531},
  year={2023}
}

@article{liu2024nonlinear,
  title={A nonlinear kernel SVM classifier via L0/1 soft-margin loss with classification performance},
  author={Liu, Ju and Huang, Ling-Wei and Shao, Yuan-Hai and Chen, Wei-Jie and Li, Chun-Na},
  journal={Journal of Computational and Applied Mathematics},
  volume={437},
  pages={115471},
  year={2024},
  publisher={Elsevier}
}

@inproceedings{li2021beyond,
  title={Beyond max-margin: Class margin equilibrium for few-shot object detection},
  author={Li, Bohao and Yang, Boyu and Liu, Chang and Liu, Feng and Ji, Rongrong and Ye, Qixiang},
  booktitle={Proceedings of the IEEE/CVF conference on computer vision and pattern recognition},
  pages={7363--7372},
  year={2021}
}

@article{khaled2023unified,
  title={Unified analysis of stochastic gradient methods for composite convex and smooth optimization},
  author={Khaled, Ahmed and Sebbouh, Othmane and Loizou, Nicolas and Gower, Robert M and Richt{\'a}rik, Peter},
  journal={Journal of Optimization Theory and Applications},
  volume={199},
  number={2},
  pages={499--540},
  year={2023},
  publisher={Springer}
}

@article{calafiore2019log,
  title={Log-sum-exp neural networks and posynomial models for convex and log-log-convex data},
  author={Calafiore, Giuseppe C and Gaubert, Stephane and Possieri, Corrado},
  journal={IEEE transactions on neural networks and learning systems},
  volume={31},
  number={3},
  pages={827--838},
  year={2019},
  publisher={IEEE}
}

@article{zhao2024adaptivity,
  title={Adaptivity and non-stationarity: Problem-dependent dynamic regret for online convex optimization},
  author={Zhao, Peng and Zhang, Yu-Jie and Zhang, Lijun and Zhou, Zhi-Hua},
  journal={Journal of Machine Learning Research},
  volume={25},
  number={98},
  pages={1--52},
  year={2024}
}

@article{townsend1971theoretical,
  title={Theoretical analysis of an alphabetic confusion matrix},
  author={Townsend, James T},
  journal={Perception \& Psychophysics},
  volume={9},
  pages={40--50},
  year={1971},
  publisher={Springer}
}

@book{boyd2004convex,
  title={Convex optimization},
  author={Boyd, Stephen and Vandenberghe, Lieven},
  year={2004},
  publisher={Cambridge university press}
}

@article{biau2016random,
  title={A random forest guided tour},
  author={Biau, G{\'e}rard and Scornet, Erwan},
  journal={Test},
  volume={25},
  pages={197--227},
  year={2016},
  publisher={Springer}
}

@inproceedings{liu2023dropout,
  title={Dropout reduces underfitting},
  author={Liu, Zhuang and Xu, Zhiqiu and Jin, Joseph and Shen, Zhiqiang and Darrell, Trevor},
  booktitle={International Conference on Machine Learning},
  pages={22233--22248},
  year={2023},
  organization={PMLR}
}

@article{wimmer2023dimensionality,
  title={Dimensionality reduced training by pruning and freezing parts of a deep neural network: a survey},
  author={Wimmer, Paul and Mehnert, Jens and Condurache, Alexandru Paul},
  journal={Artificial Intelligence Review},
  volume={56},
  number={12},
  pages={14257--14295},
  year={2023},
  publisher={Springer}
}

@article{ngo2022evolutionary,
  title={Evolutionary bagging for ensemble learning},
  author={Ngo, Giang and Beard, Rodney and Chandra, Rohitash},
  journal={Neurocomputing},
  volume={510},
  pages={1--14},
  year={2022},
  publisher={Elsevier}
}

@article{zounemat2021ensemble,
  title={Ensemble machine learning paradigms in hydrology: A review},
  author={Zounemat-Kermani, Mohammad and Batelaan, Okke and Fadaee, Marzieh and Hinkelmann, Reinhard},
  journal={Journal of Hydrology},
  volume={598},
  pages={126266},
  year={2021},
  publisher={Elsevier}
}

@ARTICLE{logsumexp,
	author={Calafiore, Giuseppe C. and Gaubert, Stephane and Possieri, Corrado},
	journal={IEEE Transactions on Neural Networks and Learning Systems}, 
	title={A Universal Approximation Result for Difference of Log-Sum-Exp Neural Networks}, 
	year={2020},
	volume={31},
	number={12},
	pages={5603-5612}}
\end{document}